\documentclass[conference]{IEEEtran}
\IEEEoverridecommandlockouts
\usepackage{cite}
\usepackage{amsmath,amssymb,amsfonts}
\usepackage{algorithmic}
\usepackage{graphicx}
\usepackage{textcomp}
\usepackage{xcolor}
\usepackage{balance}
\def\BibTeX{{\rm B\kern-.05em{\sc i\kern-.025em b}\kern-.08em
    T\kern-.1667em\lower.7ex\hbox{E}\kern-.125emX}}
\begin{document}

\title{Custom Sine Waves Are Enough for Imitation Learning of Bipedal Gaits with Different Styles
}

\author{\IEEEauthorblockN{Qi Wu$^{*}$}
\IEEEauthorblockA{\textit{Department of Mechanical Engineering} \\
\textit{Tsinghua University}\\
Beijing, China \\
wuqi19@mails.tsinghua.edu.cn \\
$^{*}$ Equal contribution}
\and
\IEEEauthorblockN{Chong Zhang$^{*}$}
\IEEEauthorblockA{\textit{Department of Precision Instrument} \\
\textit{Tsinghua University}\\
Beijing, China \\
chong-zh18@mails.tsinghua.edu.cn \\
$^{*}$ Equal contribution}
\and
\IEEEauthorblockN{Yanchen Liu}
\IEEEauthorblockA{\textit{Department of Informatics} \\
\textit{Technical University of Munich}\\
Munich, Germany \\
yanchen.liu@tum.de}
}

\maketitle

\begin{abstract}
Not until recently, robust bipedal locomotion has been achieved through reinforcement learning. However, existing implementations rely heavily on insights and efforts from human experts, which is costly for the iterative design of robot systems. Also, styles of the learned motion are strictly limited to that of the reference. In this paper, we propose a new way to learn bipedal locomotion from a simple sine wave as the reference for foot heights. With the naive human insight that the two feet should be lifted up alternatively and periodically, we experimentally demonstrate on the \textit{Cassie} robot that, a simple reward function is able to make the robot learn to walk end-to-end and efficiently without any explicit knowledge of the model. With custom sine waves, the learned gait pattern can also have customized styles. Codes are released at github.com/WooQi57/sin-cassie-rl.
\end{abstract}


\section{Introduction}
\label{sec:intro}
Bipedal robot locomotion has long been a challenging task where classical methods typically rely on simplified models \cite{kajita2003biped} \cite{vukobratovic2004zero} \cite{englsberger2011bipedal} \cite{reher2016realizing} \cite{yang2022bayesian}, consequently limiting the agility of robots with small regions of attraction for possible motions. 
Recently, emerging model-free methods such as reinforcement learning (RL) have shown advantages over traditional methods by fully exploring the dynamic of robots and implicitly modeled information of the environments \cite{li2021reinforcement}. Through trials and errors, robots can explore feasible policies by their own.

However, such a data-driven way also requires well-designed reward functions and lengthy training \cite{rodriguez2021deepwalk}, which often take massive efforts for tuning. Therefore, imitation learning has become popular because of the high data efficiency achieved by guiding the robot with references \cite{lee2010data}. Yet the acquisition of references is not trivial. To guarantee the quality of motions to imitate, the references also require insights and efforts from human experts. Common practices for generating references are manually-tuned controllers \cite{xie2018feedback} and costly motion captures (mocaps) \cite{peng2017deeploco}.

\begin{figure}[t]
  \centering
    \includegraphics[width=87mm]{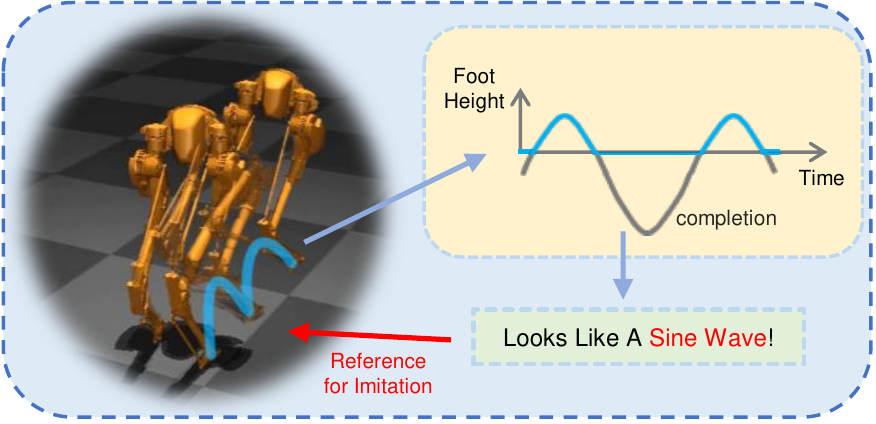}
    \caption{
       For bipedal walking, the foot height curve w.r.t. time looks like a sine wave if completed. This leads to our simple idea that, a sine wave is enough to serve as the reference for imitation learning of bipedal locomotion.
    } 
    \label{fig:idea}
  \vspace{-1em}
\end{figure}

Therefore, it makes sense to find ways that can reduce the cost and efforts for reference generation. And in this paper, we find that even a sine wave is enough to generate the references for bipedal locomotion, with quite simple system configurations and reward functions. Specifically, we start from a simple idea that walking means lifting the two feet alternatively for bipedal robots, and the foot heights can be approximated by the positive part of a sine wave, as is illustrated in Fig. \ref{fig:idea}. 

Despite such a simple idea, to our best knowledge, no existing work has achieved data-efficient imitation learning from such easy-to-generate references for bipedal locomotion. In other words, the references generated in our paper are currently the simplest references that can be used for imitation learning of bipedal locomotion, with no need of advanced knowledge such as Bezier curves or kinematics. Moreover, with such simple but effective references, we can not only achieve bipedal locomotion in different directions, but also generate gait patterns of different styles.

\section{Related Works}
\label{sec:related}

Regarding RL, the model-free algorithm Proximal Policy Optimization (PPO) \cite{schulman2017proximal} has become a routine for bipedal locomotion learning. The learning strategies of existing works can be split into two categories: 1) learning without references, and 2) learning from references. 

\subsection{Learning without References}
\label{subsec: w/o ref}
Policies can be trained in simulation directly without a reference, as demonstrated by many works \cite{rodriguez2021deepwalk} \cite{siekmann2021sim} \cite{siekmann2021blind}. However, these works all require lengthy training processes that can take several dozens of hours, and well-designed reward functions that would conceivably take laborious tuning. Such methods are unacceptable especially when we are trying to verify the agility of a robot that is undergoing iterative design.

Although there are other works claiming fast training and successful sim2real on legged robots \cite{makoviychuk2021isaac} \cite{rudin2021learning}, we found it difficult to deploy the learned policies on real bipedal robots, because the fast learning is achieved by massive parallel training based on GPU, and the dynamics in simulation are not accurately modeled. Although quadrupedal robots can tolerate such inaccuracy, we found that bipedal robots are much more vulnerable to unreliable dynamics.

\subsection{Learning from References}
\label{subsec: w/ ref}

Another common practice is to imitate motions. \cite{peng2017deeploco} \cite{peng2018deepmimic} and \cite{peng2020learning} used an imitation learning framework which enables bipeds to learn multiple tasks in simulation by imitating human mocaps. While mocaps are difficult to obtain, model-based controllers can generate the motion for reference. \cite{xie2018feedback} \cite{xie2019iterative} achieved imitation learning with sim2real from the motion generated by a manually-tuned controller. \cite{green2021learning} \cite{duan2021learning} used the actuated spring loaded inverted pendulum (ASLIP) model as a reduced-order model to generate the expert controllers. Another choice for the model-based controller is hybrid zero dynamics (HZD) \cite{westervelt2018feedback} \cite{nguyen2020dynamic}. In \cite{li2021reinforcement}, an HZD library with $>1$k controllers is used for imitation learning of omnidirectional walking. 

Despite the recent success of imitation learning from these references, it takes time and effort to generate them. Also, the learned gait pattern always follows the style of the reference, which can be difficult to tune and sometimes cannot support omnidirectional locomotion.

Besides, sim2real transfer is a problem, but as suggested in \cite{siekmann2021sim} and \cite{siekmann2021blind}, sim2real and skill learning can be separately treated. With appropriate domain randomization \cite{tobin2017domain}, learning the policy in simulation is enough for sim2real transfer, even when the task is blind stair traversal \cite{siekmann2021blind}. Following this idea, we would like to focus on skill learning in this paper without the consideration of sim2real, so that we can present our work in a more straightforward but powerful way.

\subsection{Motivation}
\label{subsec: motiv}
Based on existing works, we believe imitation learning is still necessary in many cases, and so are the references. However, with existing works using mocaps and manually-tuned model-based controllers to get the references, we are motivated to find a simple representation that can hint the learning without laborious work. Also, we want to enable the robot to walk with different styles instead of being limited by the reference.

To this end, we propose to use sine waves as the reference in this paper. We show that, with easy design of the reward function and end-to-end training, a simple sine wave is enough for the reference in the imitation learning formulation. Hopefully, our method can make RL less an accessory to existing controllers or an exhausting big project, but more a flexible tool to inspect the agility of a bipedal robot.

\section{Methodology}
\label{sec:method}
\subsection{System Overview}
\label{subsec:system}
\begin{figure}[t]
    \centering
    \includegraphics[width=87mm]{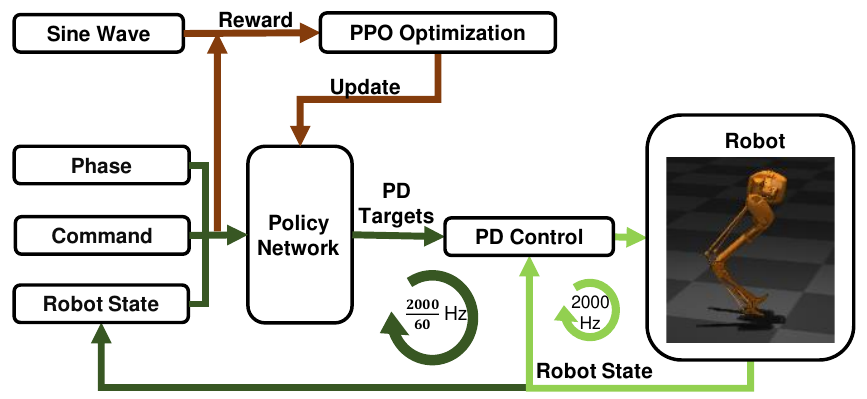}
    \caption{
      System Overview. The red arrows indicate the optimization-related parts, the dark green arrows indicate the inputs and the outputs of the policy network, and the light green arrows indicate the low-level control in the simulation environment.
    } 
    \label{fig:sys overview}
\vspace{-1em}
\end{figure}

In a typical RL framework, an agent learns a task through trials and errors in the environment. The task contains ${S,A,p,\gamma,r}$, where $S$ is the state space, $A$ means possible actions, $\gamma\in[0,1]$ is the discount factor, $p$ indicates the dynamics and $r$ is the reward function that gives out a reward at each state. The process is to learn a policy $\pi$ that outputs the action at a given state which maximizes the return value.  

In this paper, we implemented our method on the 20-DOF bipedal robot Cassie in the MuJoCo simulator \cite{todorov2012mujoco} with cassie-mujoco-sim environment \cite{cassie-mujoco-sim}. The system is illustrated in Fig. \ref{fig:sys overview}. The simulation runs at 2000 Hz. The policy network outputs the target joint positions per 0.03 s, and the target positions are passed to a low-level PD controller that runs at 2000 Hz. Here we adopt the positional control for better learning efficiency and performance according to \cite{peng2017learning}. 

The inputs of the policy network consist of the position $q_j$ and the velocity $\dot{q_j}$ of all 14 joints, the orientation of pelvis $o_p$ in the form of quaternion, the angular velocity of the pelvis $\Omega_p$, the translational velocity of the pelvis $v_p$, the height of the pelvis $h_p$, the phase vector $[\sin(\frac{2\pi}{T}t+\phi_0),\cos(\frac{2\pi}{T}t+\phi_0)]$, and the velocity command $v_c=[v_x,v_y]$. The period $T$ in the phase vector corresponds to the frequency we want for bipedal walking. 

\subsection{Reward}
The reward function is designed to encourage the robot to lift its feet periodically while catching up with the given velocity command. The reward is calculated at each step as
\begin{equation}
    r_t = 0.5 r_t^I + 0.5 r_t^P + r_t^R + r_t^{T},
    \label{eq:rew form}
\end{equation}
where $r_t^I$ is the \textbf{normalized} imitation reward term encouraging imitation, $r_t^P$ is the performance reward term encouraging velocity tracking and orientation control, $r_t^R$ is the regularization reward term encouraging more realistic motion, and $r_t^T$ is the termination reward term discouraging falling and accumulative tracking errors.

The \textbf{normalized} imitation reward term incentivizes the robot to lift its feet as the sine wave reference. It is derived from the \textbf{nominal} imitation reward term,
\begin{equation}
    r_t^{I*} = \exp{(-\frac{1}{0.05^2}\sum_{i=1}^2(h_{\rm{ref}_i}-h_{\rm{foot}_i})^2)},
    \label{eq:imi}
\end{equation}
where the differences between the reference heights and the current heights of both feet are discouraged. The normalization of $r_t^{I*}$ is vital, which will be explained in Sec. \ref{subsec:norm}. It is defined as
\begin{equation}
    r_t^{I} = \frac{r_t^{I*}-B_{\rm lower}}{B_{\rm upper}-B_{\rm lower}},
    \label{eq:norm}
\end{equation}
where $B_{\rm upper}$ is the upper bound for $r_t^{I*}$ and we casually assign $B_{\rm upper}=1$. $B_{\rm lower}$ is the lower bound for $r_t^{I*}$ \textit{if the robot is pursuing higher imitation reward}, and we casually assign it as $B_{\rm lower}=0.4$, slightly above the $r_t^{I*}$ value for a random policy. In other words, the \textbf{normalized} imitation reward term becomes a penalty for survival if the agent does not pursue imitation. We find the normalization critical in our implementation, and further explanations are in Sec. \ref{subsec:norm}.

$r_t^P$ indicates the reward for performance. It's formulated as
\begin{equation}
    r_t^P = 0.75\exp{(-p_v)}+0.25\exp{(-p_o)},
    \label{eq:performance}
\end{equation}
where
\begin{equation}
    p_v = \frac{\lVert [v_{p,x},v_{p,y}]-v_c \rVert^2}{
    \max \left(0.1^2,0.5\lVert v_c \rVert ^2\right) },
    \label{eq:vel pelvis}
\end{equation}
which focuses on velocity tracking for the pelvis, and
\begin{equation}
    p_o = \frac{\sin^2(0.5\langle o_p, o_u \rangle)}{0.1},
    \label{eq:ori pelvis}
\end{equation}
which focuses on orientation control of the pelvis. Here $\langle \cdot, \cdot \rangle$ is the angle of two orientations and $o_u$ the orientation of standing upright facing $+x$ direction.

The last two items $r_t^R$ and $r_t^T$ in (\ref{eq:rew form}) are the penalties for shin springs and termination. Specifically,
\begin{equation}
r_t^R = 0.1 \exp (-\frac{q_{\rm leftShin}^2+q_{\rm rightShin}^2}{0.001}),
    \label{eq:shin}
\end{equation}
where the joint angles of the two shins are penalized.
\begin{equation}
r_t^T = -10 \text{ if }\tt{terminate}\text{ else } 0,
    \label{eq:term}
\end{equation}
where termination is activated if 1) the height of pelvis is lower than 0.6, or 2) the height of pelvis is higher than 1.2, or 3) the position of the robot is too far away from the position it should be at according the velocity command. To be specific, we define the third condition as 
\begin{equation}
    \lVert [x_p,y_p,h_p] - [x_t, y_t, z_t] \rVert
    \le 0.6+\lVert v_c\rVert,
    \label{eq:3rd_cond}
\end{equation}
where $[x_p,y_p,h_p]$ is the position of the pelvis, and $[x_t, y_t, z_t]$ is the position calculated by integrating the velocity command. This condition enforces accurate tracking with limited cumulative errors.

It is worth mentioning that, all of the weights in our reward function are set from experience without any special efforts of fine-tuning. The normalization is non-trivial, because no existing work has achieved imitation learning from such simple references and configurations, and we attribute this to the fact that no existing work to our knowledge penalizes the robot for survival.

\subsection{Reference}

\begin{figure}[t]
  \centering
    \includegraphics[width=87mm]{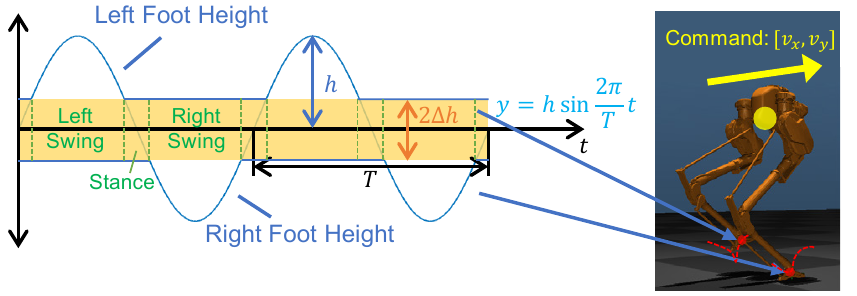}
    \caption{
       Generated gait references for the robot's feet during the tracking of a velocity command.
    } 
    \label{fig:ref}
  \vspace{-1em}
\end{figure}

References are generated from a simple sine wave, as is shown in Fig. \ref{fig:ref} and defined below:
\begin{equation}
    h_{\rm{ref}_1}=h_{\rm{left}} = \max(0, h\sin(\frac{2\pi}{T}t+\phi_0)-\Delta h),
    \label{eq:ref left}
\end{equation}
\begin{equation}
    h_{\rm{ref}_2}=h_{\rm{right}} = \max(0, h\sin(\frac{2\pi}{T}t+\phi_0+\pi)-\Delta h).
    \label{eq:ref right}
\end{equation}

The heights of both feet are the positive part of a sine wave with a bias. The maximum height of a foot is $h-\Delta h$, and the phase of the sine wave is consistent with that in the phase vector mentioned in Sec. \ref{subsec:system}. The $-\Delta h$ term used in (\ref{eq:ref left}) and (\ref{eq:ref right}) are to leave a time span for double-support stance, which takes up $4\arcsin(\frac{\Delta h}{h})$ of the whole $2\pi$ period. As is shown in Fig. \ref{fig:ref}, the reference simply gives the incentive for lifting the corresponding foot at a certain phase. 

Different values for $h$, $\Delta h$, and $T$ can describe different walking styles. Larger $h-\Delta h$ values can lead to larger foot clearance, which affects the whole-body posture. Larger $\Delta h/h$ values can lead to longer double-support stance, which also brings faster feet up and down. Smaller $T$ values can lead to higher frequencies, which makes the gaits more hurried. 

\begin{figure*}[t]
    \centering
    \includegraphics[width=170mm]{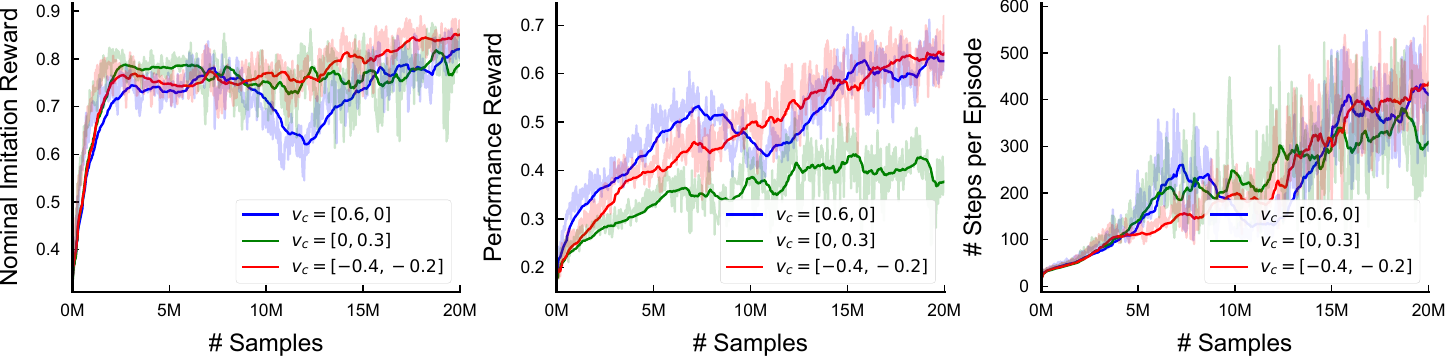}
    \caption{
       The learning curves for different velocity commands. The solid lines are smoothed by exponential moving average for the values in shadow. The imitation rewards all increase sharply at first indicating successful imitation, and the performance rewards gradually go higher with the number of survival steps. More than 300 survival steps, i.e., 9 s, can empirically promise successful learning of the tracking, and further learning is marginal.
    } 
    \label{fig:command}
  
\vspace{-1em}
\end{figure*}

\section{Implementation}
\label{sec:implementation}
\subsection{Configurations}
\label{subsec:config}
We used stable-baselines3 \cite{stable-baselines3} to implement our PPO algorithm with GAE \cite{Schulmanetal_ICLR2016}. Two hidden layers with 512 ReLU units are used to for the policy and the value function. With a mini-batch size of 128, we optimized the policy per 256 steps in 16 parallel environments, i.e., per 4096 samples. Other hyperparameters go with the default values in stable-baselines3, where the learning rate is 3e-4, the $\gamma$ value is  0.99, and the $\lambda$ value for GAE is 0.95.

We initialize the robot as standing upright facing $+x$ direction with no speed. The initial phase $\phi_0$ is randomly set as $0$ or $\pi$, which corresponds to the double-support stance, and the first step can be $50\%$ chance the left foot, $50\%$ chance the right foot.

Remarkably, if we predefine a pose for $\phi_0=0$ with the left foot behind the right foot, and the mirror pose for $\phi_0=\pi$ with the right foot behind the left foot, the learning efficiency can be greatly improved by $\sim 40\%$. This is a natural way in imitation learning to reduce unnecessary exploration that goes away from the motion we want. However, we are not sure whether one pose can work for different velocities in different directions, and such an implementation is not easily reproducible, so we still choose to initialize the robot with the default pose.

\subsection{Velocity Tracking}
\label{subsec:velo tracking}

\begin{figure}[t]
  \centering
    \includegraphics[width=87mm]{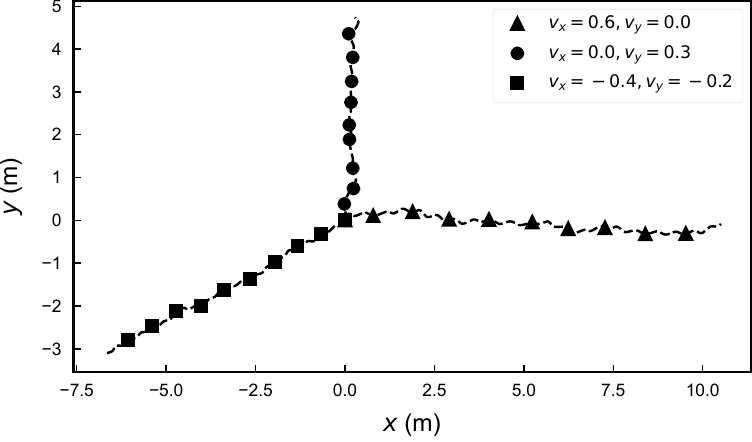}
    \caption{
       The positions of the pelvis for tracking of different velocity commands. Marks are at intervals of 1.8 s.
    } 
    \label{fig:tracking}
  \vspace{-1em}
\end{figure}

\begin{figure}[t]
  \centering
    \includegraphics[width=87mm]{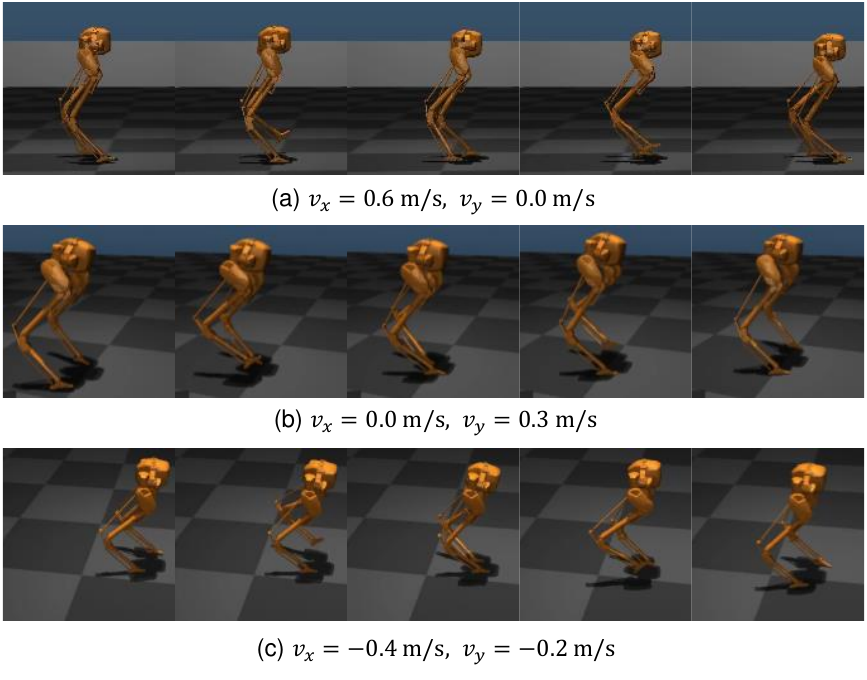}
    \caption{
       The robot can follow the same reference while tracking different velocities in different directions.
    } 
    \label{fig:poses1}
  \vspace{-1em}
\end{figure}

To show that our simple reference from a sine wave is enough for learning to walk in different directions at different speeds, we tried three different velocity commands: 
\begin{enumerate}
    \item $v_x=0.6 {\rm m/s}, v_y=0{\rm m/s}$;
    \item $v_x=0 {\rm m/s}, v_y=0.3{\rm m/s}$;
    \item $v_x= -0.4 {\rm m/s}, v_y= -0.2 {\rm m/s}$.
\end{enumerate}
We casually assigned $h-\Delta h=0.12$ m and $\Delta h/h=0.2$, with $T=28$ timesteps, i.e., 0.84 s, to generate the same reference for all 3 commands. 

The learning curves are shown in Fig. \ref{fig:command}, where we choose to display the \textbf{nominal} imitation reward $r_t^I$, the performance reward $r_t^P$, and the survival steps per episode for easy analysis. We do not show the total reward curves because the termination condition of accurate tracking in (\ref{eq:3rd_cond}) makes the survival steps a better indicator for the learning progress. 

In Fig. \ref{fig:tracking}, we showcase the positions of the pelvis when the robot is tracking different velocities in different directions. Thanks to the accurate tracking condition in  (\ref{eq:3rd_cond}), there is little cumulative error although the robot starts from zero speed and we do not provide any information about the $x-y$ positions.

\begin{figure}[t]
  \centering
    \includegraphics[width=80mm]{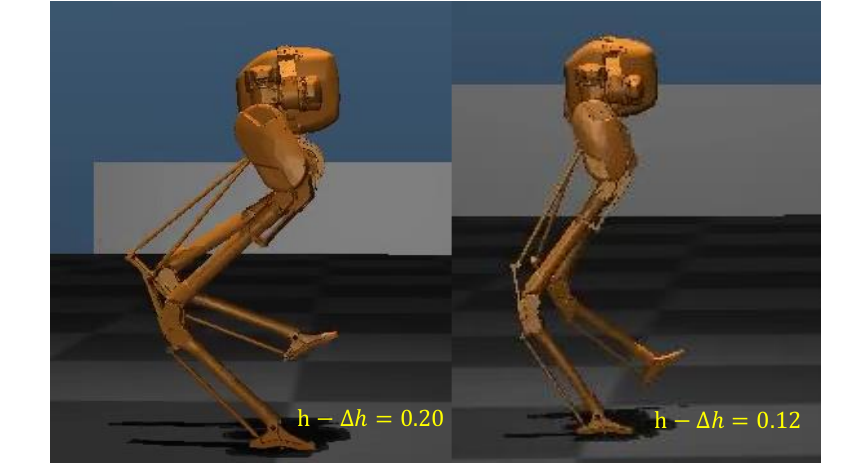}
    \caption{
       Different $h-\Delta h$ values bring not only different foot clearance but also different whole-body posture. Snapshots were taken when the left foot was at its maximum height.
    } 
    \label{fig:compare height}
  \vspace{-1em}
\end{figure}

\begin{figure*}[t]
  \centering
    \includegraphics[width=170mm]{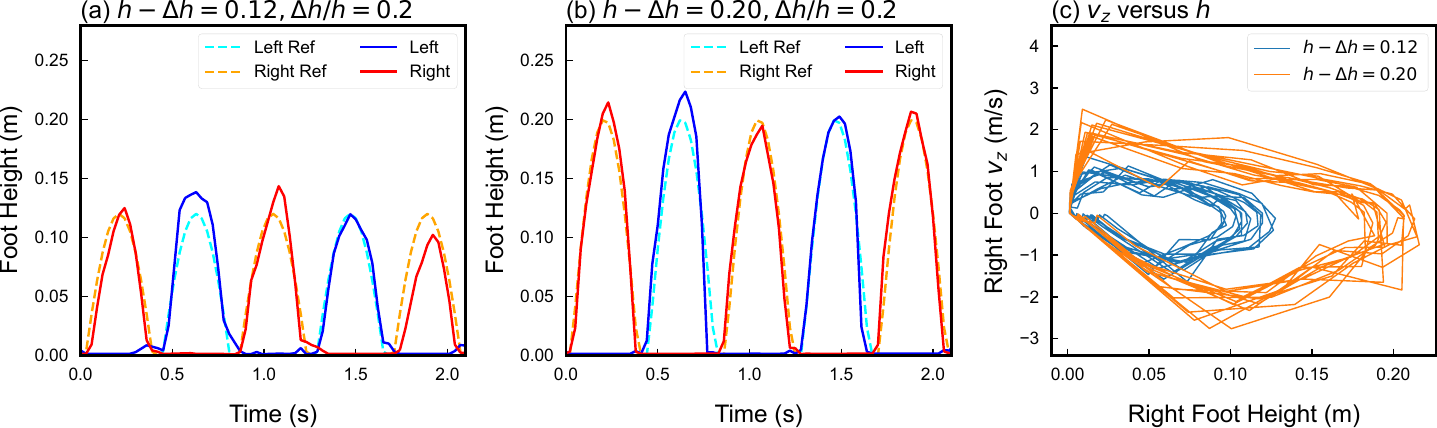}
    \caption{
       Foot height curves for different $h-\Delta h$ values. The robot learned to lift its feet to different heights. (a) is with $h-\Delta h=0.12$ m and $\Delta h/h=0.2$. The figure shows the references and the robot's foot heights. (b) changes the max height to 0.20m. (c) shows the relationship between the height of one foot and the foot's vertical velocity.  
    } 
    \label{fig:footheight}
\vspace{-1em}
\end{figure*}

\begin{figure*}[t]
    \centering
    \includegraphics[width=170mm]{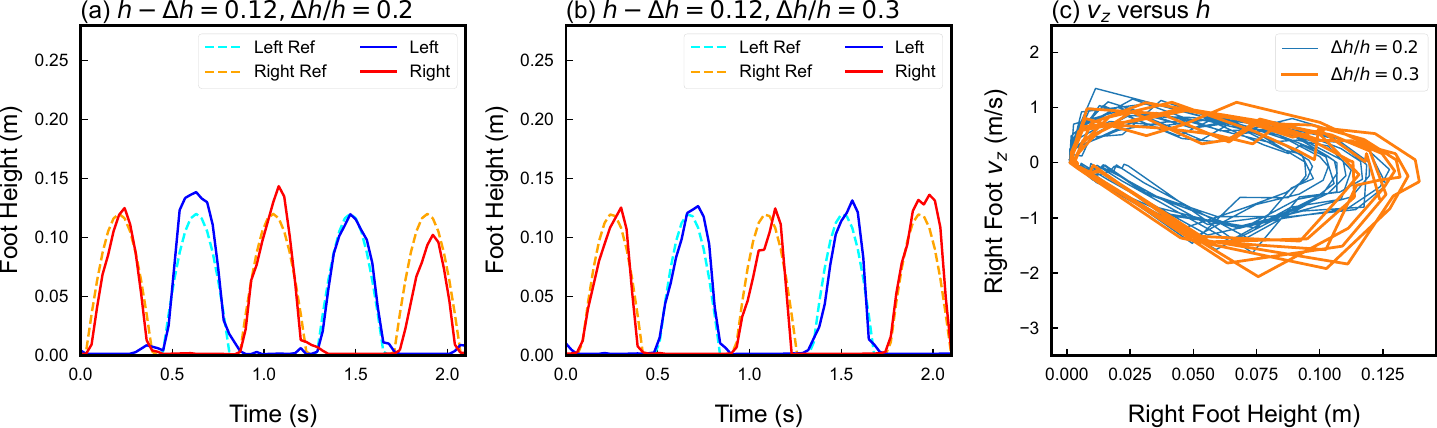}
    \caption{
       Foot height curves for different double-support spans. The maximum feet heights are 0.12 m and the period is $T=0.84$ s. (a) is with $\Delta h/h=0.2$, (b) is with $\Delta h/h=0.3$, and (c) shows that a longer double-support span brings faster landing.
    } 
    \label{fig:footheight2}
\vspace{-1em}
\end{figure*}

\begin{figure*}[t]
    \centering
    \includegraphics[width=170mm]{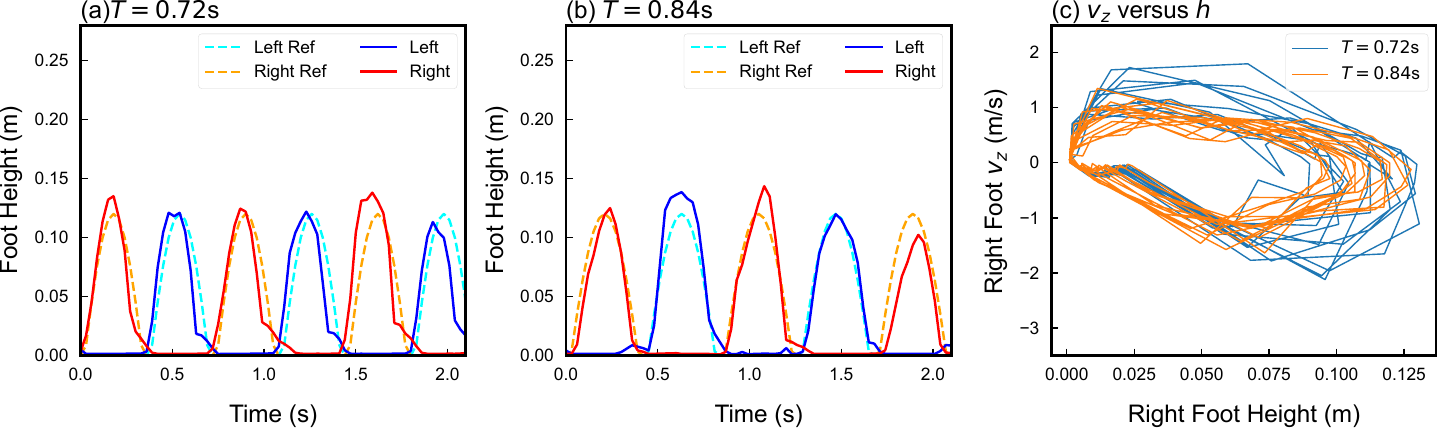}
    \caption{
       Foot height curves for different gait frequencies. The maximum feet heights are 0.12 m and $\frac{\Delta h}{h}=0.2$. (a) is with $T=0.72$ s, (b) is with $T=0.84$ s, and (c) shows that a higher frequency leads to faster feet up and down.
    } 
    \label{fig:footheight3}
  
\vspace{-1em}
\end{figure*}

As is shown in Fig. \ref{fig:poses1}, with our method, one reference can work for multiple velocities in different directions, because we only provide information for foot height control. In contrast, controller-based references can typically support only one velocity, and multiple velocities require multiple references \cite{li2021reinforcement} \cite{xie2018feedback}. Some controllers can only support locomotion in one direction \cite{green2021learning} \cite{duan2021learning}.

\subsection{Different Styles}
\label{subsec:diff styles}

To show that our method supports learning of gaits with different styles, we showcase in this subsection how different values of $h$, $\Delta h$, and $T$ can influence the style. Specifically, we took the same velocity command $v_x=0.6 {\rm m/s}, v_y=0{\rm m/s}$, and tried three comparisons: 1) changing $h-\Delta h$ for foot clearance, 2) changing $\Delta h/h$ for double-support spans, and 3) changing $T$ for different frequencies.

\subsubsection{Changing $h-\Delta h$}
Here we took two $h-\Delta h$ values 0.12 m and 0.20 m as a comparison, with $\Delta h/h=0.2$ and $T=28$ timesteps, i.e., 0.84 s. Snapshots are in Fig. \ref{fig:compare height}, and foot height curves are shown in Fig. \ref{fig:footheight}.

\subsubsection{Changing $\Delta h/h$}
With $h-\Delta h=0.12$ m and $T=0.84$ s, we tried two different values for $\Delta h/h$, 0.2 and 0.3. The results are shown in Fig. \ref{fig:footheight2}, indicating that a longer double-support span brings faster landing.

\begin{figure*}[t]
    \centering
    \includegraphics[width=170mm]{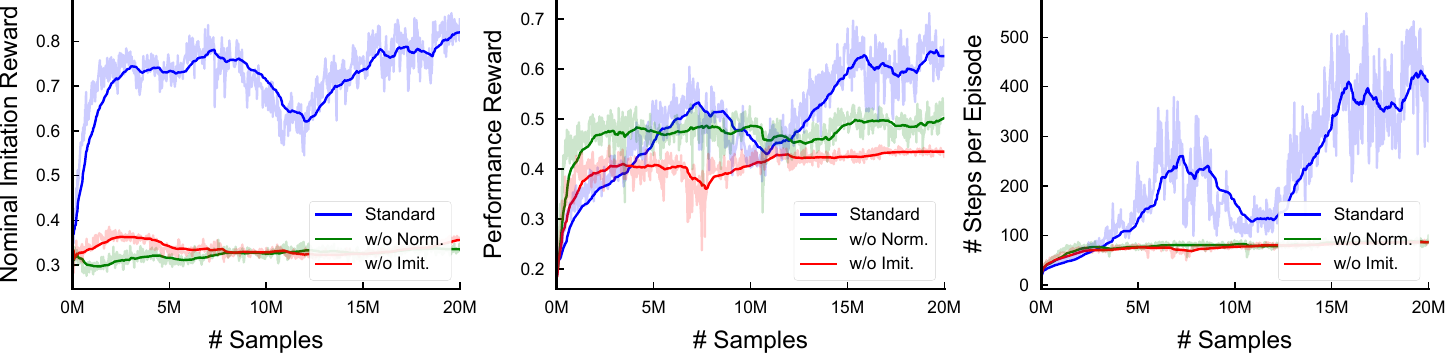}
    \caption{
       Learning curves for ablation studies of imitation and normalization. The results are for $v_x=0.6 {\rm m/s}, v_y=0 {\rm m/s}, h-\Delta h = 0.12 {\rm m}, \Delta h/h = 0.2, T =0.84{\rm s}$. 
    } 
    \label{fig:ablation}
  
\vspace{-1em}
\end{figure*}

\subsubsection{Changing $T$}
With $h-\Delta h=0.12$ m and $\Delta h/h=0.2$, we tried different periods $T=0.72$ s and $T=0.84$ s. As is shown in Fig. \ref{fig:footheight3}, a higher frequency leads to a hurried gait pattern with faster feet up and down.

\section{Discussion}
\label{sec:discuss}

\subsection{Imitation}
\label{subsec:discuss imit}
We successfully achieved imitation learning of bipedal locomotion from the very simple references. The results show that a hint for lifting feet is enough for learning. The generation of references in our paper is much easier than designing multiple model-based controllers or collecting mocaps. On the other hand, compared with learning without references, our method can expedite the learning process and reduce laborious tuning. In Fig. \ref{fig:ablation}, it is shown that the robot cannot learn to walk using our simple configurations without imitation learning from the reference, i.e., $r_t=\mathbf{0}r_t^I +\mathbf{1}r_t^P+r_t^R+r_t^T$.

\subsection{Normalization}
\label{subsec:norm}
The normalization in (\ref{eq:norm}) is necessary so that the agent is forced to imitate before trying to survive more time. In this way, the imitation rewards can quickly go up in the very first several episodes, indicating that the robot is trying to lift its feet. Otherwise, the longer survival time can be a penalty. The curves in Fig. \ref{fig:ablation} show that the robot cannot learn to work without the normalization of the imitation reward term, i.e., $r_t = 0.5 \mathbf{r_t^{I*}} + 0.5 r_t^P+r_t^R+r_t^T$.

The idea in this paper is quite simple, but no existing work to our knowledge has achieved imitation learning from such a easily generated reference. We believe this is because, no existing work has come to the idea of penalizing the "living" robot. In most of the cases, people just do nothing or even give positive rewards for the "living" robot, but rather we penalize it by normalization if imitation is not strongly pursued.

\subsection{Different Styles}
\label{subsec:discuss styles}
In our implementation, we show that the gait patterns can be changed with different parameters in the sine wave. In contrast, existing works always limit the style of the learned motion to that of the references. In brief, our method can allow more space to customize the gait pattern.

\subsection{Learning Efficiency}
In this paper, we use the very simple references, the very simple configurations, and the very simple reward functions to train the policies. Each policy took $\sim 5$ hours to train on an ordinary desktop machine. Still, the training time can be decreased to $< 3$ hours if:
\begin{enumerate}
    \item Appropriate poses for initialization can be provided, as is mentioned in Sec. \ref{subsec:config}. According to \cite{zhang2022accessibility}, initial state distributions can greatly affect the data efficiency.
    \item Accurate tracking is not enforced. This can further greatly improve the learning efficiency, but can lead to cumulative tracking errors. For example, in \cite{li2021reinforcement}, there is a large gap between the velocity command and the real velocity.
\end{enumerate}

\section{Conclusion and Future Work}
\label{sec:conclusion}

In this paper, we propose to help the bipedal robot learn to walk at different velocities, in different directions, and with different styles. This is achieved by imitation learning from the very simple references generated by custom sine waves. We also analyze in our paper how and why it can work with our very simple configurations and reward functions. Hopefully, this work can free reinforcement learning from laborious tuning for either reward functions and learning strategies, or model-based controllers to imitate. With the high learning efficiency, the proposed method may also be used to verify the agility of a bipedal robot that is undergoing iterative design.

Future works may be focused on validation and extension of the proposed method on other robot platforms, and the sim2real transfer of the learned policy. Also, we are expecting to express all periodical motions with combination of sine waves, and further extend our method to diverse tasks.

\bibliographystyle{IEEEtran}
\balance
\bibliography{IEEEabrv,ref}

\end{document}